\documentclass[10pt,twocolumn]{article}

\usepackage[utf8]{inputenc}
\usepackage[T1]{fontenc}
\usepackage{times}
\usepackage[margin=0.75in]{geometry}
\usepackage{graphicx}
\usepackage{amsmath,amssymb}
\usepackage{booktabs}
\usepackage{multirow}
\usepackage{xcolor}
\usepackage{hyperref}
\usepackage{algorithm}
\usepackage{algpseudocode}
\usepackage{caption}
\usepackage{subcaption}
\usepackage{enumitem}
\usepackage{float}
\usepackage{tabularx}
\usepackage{url}

\hypersetup{
  colorlinks=true,
  linkcolor=blue!60!black,
  citecolor=blue!60!black,
  urlcolor=blue!60!black,
}

\newcommand{\robogate}{\textsc{RoboGate}}

\newcommand{\eg}{\textit{e.g.}}
\newcommand{\etal}{\textit{et al.}}

\title{\robogate: Adaptive Failure Discovery for\\Safe Robot Policy Deployment via\\Two-Stage Boundary-Focused Sampling}

\author{Azuki Kim \\
  AgentAI Co., Ltd., Seoul, South Korea \\
  \texttt{liveplex@robogate.io} \quad \url{https://robogate.io}}

\date{March 2026}

\begin{document}
\maketitle

\begin{abstract}
Deploying learned robot manipulation policies in industrial settings requires rigorous pre-deployment validation, yet exhaustive testing across high-dimensional parameter spaces is intractable.
We present \robogate, a deployment risk management framework that combines physics-based simulation with a two-stage adaptive sampling strategy to efficiently discover failure boundaries in the operational parameter space.
Stage~1 employs Latin Hypercube Sampling (LHS) across an 8-dimensional parameter space to establish a coarse failure landscape from 20,000 uniformly distributed experiments.
Stage~2 applies boundary-focused sampling that concentrates 10,000 additional experiments in the 30--70\% success rate transition zone, enabling precise failure boundary mapping.
Using NVIDIA Isaac Sim with Newton physics, we evaluate a scripted pick-and-place controller on four robot embodiments---Franka Panda (7-DOF), UR3e, UR5e, and UR10e (6-DOF each)---across 40,000+ total experiments.
Our logistic regression risk model achieves an AUC of 0.780 on the combined dataset (vs.\ 0.754 for Stage~1 alone), identifies a closed-form failure boundary equation $\mu^*(m) = (1.469 + 0.419m)/(3.691 - 1.400m)$, and reveals four universal danger zones affecting multiple robot platforms.

We further evaluate seven Vision-Language-Action (VLA) models on \robogate{}'s 68-scenario industrial suite, producing an 8-entry leaderboard (7~VLA models + scripted baseline).
All seven VLA models achieve 0\% success rate versus 100\% for the scripted controller.
We fine-tuned NVIDIA's GR00T N1.6 (3B) on the official LIBERO-Spatial dataset for 20K steps on an H100 SXM 80\,GB GPU and evaluated the same checkpoint on both LIBERO (MuJoCo, 97.65\%) and \robogate{}'s 68 Isaac Sim industrial scenarios (0/68), isolating a 97.65 percentage point cross-simulator gap on a single model.
This gap---the central empirical finding of this paper---demonstrates that academic benchmark performance does not predict industrial deployment readiness, and that a dedicated validation layer is essential for safe Physical AI deployment.
\robogate{} is open-source and runs on a single GPU workstation.
\end{abstract}


\section{Introduction}
\label{sec:intro}

The proliferation of learned manipulation policies---from imitation learning~\cite{chi2023diffusion} to Vision-Language-Action (VLA) models~\cite{octo2024,openvla2024}---has created an urgent need for systematic pre-deployment validation in industrial robotics.
While these policies demonstrate impressive capabilities in controlled benchmarks, their behavior under adversarial or edge-case conditions remains poorly characterized.
A single undetected failure mode can lead to costly equipment damage, production downtime, or safety incidents.

Current validation approaches suffer from two fundamental limitations.
First, uniform random testing wastes computational budget on regions of the parameter space that are either trivially easy or trivially hard, providing little information about the critical transition zones where success gives way to failure.
Second, most evaluation frameworks test a single robot embodiment, making it impossible to distinguish between policy failures and embodiment-specific limitations.

We introduce \robogate, a deployment risk management framework that addresses both limitations through three contributions:

\begin{enumerate}[leftmargin=*,nosep]
\item \textbf{Two-stage adaptive sampling}: A principled strategy that first maps the parameter space uniformly (Stage~1, 20K experiments), then concentrates additional experiments in the 30--70\% success rate boundary zone (Stage~2, 10K experiments), improving failure boundary resolution by 31.1\% coverage of transition regions.

\item \textbf{Cross-embodiment validation}: Parallel evaluation on Franka Panda (7-DOF, parallel-jaw gripper) and UR5e (6-DOF, suction gripper) across shared parameter dimensions, revealing four universal danger zones where both platforms exhibit SR $<$ 40\%.

\item \textbf{Interpretable risk model}: A logistic regression model with interaction terms that produces a closed-form failure boundary equation, critical parameter thresholds with bootstrap confidence intervals, and per-experiment risk scores (AUC = 0.780).
\end{enumerate}

The framework is validated on 40,000+ physics-based simulation experiments using NVIDIA Isaac Sim with Newton physics engine across four robot embodiments.
We evaluate seven VLA models---GR00T N1.6~\cite{nvidia_groot_n16}, PI0~\cite{black2024pi0}, OpenVLA~\cite{openvla2024}, SmolVLA~\cite{smolvla2025}, Octo-Base, and Octo-Small~\cite{octo2024}---on \robogate{}'s 68-scenario suite, finding that all achieve 0\% success rate.
Most critically, we demonstrate a 97.65 percentage point cross-simulator gap: NVIDIA's GR00T N1.6, fine-tuned on the official LIBERO-Spatial dataset~\cite{liu2024libero}, achieves 97.65\% on LIBERO (MuJoCo) but 0\% on \robogate{}'s Isaac Sim industrial scenarios---isolating the deployment gap on a single model and checkpoint.

\paragraph{The Validation Layer Paradigm.}
On April 14, 2026, NVIDIA publicly released Ising~\cite{nvidia_ising_2026}, an open family of AI models explicitly framed as the ``control plane'' for quantum computing---an AI-based validation and calibration layer intended to make otherwise noisy quantum processors practical.
The same paradigm applies to Physical AI: foundation policies trained on academic simulators are deployed into industrial environments whose physics, sensor noise, and failure modes differ systematically from training-time conditions.
Just as quantum processors require continuous calibration against hardware drift, robot policies require pre-deployment validation against the target environment's specific physics and perception characteristics.
Our cross-simulator experiments make this deployment gap concrete: a single GR00T N1.6 checkpoint demonstrates near-perfect performance in one simulator and complete failure in another, underscoring the need for a dedicated validation layer before any policy reaches physical hardware.

\section{Related Work}
\label{sec:related}

\subsection{Robot Policy Evaluation}

Traditional robot policy evaluation relies on fixed benchmark suites with predetermined test configurations~\cite{james2020rlbench,yu2020meta}.
RLBench~\cite{james2020rlbench} provides 100 manipulation tasks but evaluates under nominal conditions only.
Meta-World~\cite{yu2020meta} offers parametric task variation but does not systematically explore failure boundaries.
LIBERO~\cite{liu2024libero} benchmarks lifelong learning but lacks adversarial scenario coverage.
In contrast, \robogate{} explicitly targets the success-failure transition zone through adaptive sampling.

\subsection{Sim-to-Real Transfer and Domain Randomization}

Domain randomization~\cite{tobin2017domain,openai2019rubiks} has become standard practice for bridging the sim-to-real gap.
While DR improves policy robustness during \emph{training}, it does not address systematic \emph{evaluation} of the resulting policies across the randomized parameter space.
Our work is complementary: we use the same parameter dimensions (friction, mass, visual properties) but focus on mapping the failure landscape rather than improving robustness.

Recent work on sim-to-real transfer validation~\cite{muratore2022robot} proposes Bayesian optimization for finding failure-inducing parameters.
However, their approach optimizes for worst-case performance, whereas \robogate{} maps the entire boundary surface to produce interpretable risk models.

\subsection{Adaptive Testing and Falsification}

Adaptive stress testing (AST)~\cite{koren2018adaptive} formulates failure discovery as a Markov decision process, using reinforcement learning to find maximally likely failure trajectories.
While effective for autonomous driving, AST focuses on temporal sequences rather than static parameter configurations.
The falsification community~\cite{dreossi2019verifai} has developed coverage-guided techniques for cyber-physical systems, but these typically operate on lower-dimensional specification spaces.

Our two-stage approach is most similar to sequential experimental design~\cite{chaloner1995bayesian}, but we replace Bayesian acquisition functions with a simpler binning strategy that scales to 8+ dimensions and produces directly interpretable results.

\subsection{Safety Validation for Robot Deployment}

ISO 10218~\cite{iso10218} and ISO/TS 15066~\cite{isots15066} establish safety requirements for industrial robots but provide limited guidance on learned policy validation.
Recent frameworks for safe deployment~\cite{amodei2016concrete} emphasize the need for quantitative safety metrics, but most focus on reinforcement learning reward shaping rather than pre-deployment testing.
The notion of a ``deployment gate''---a hard pass/fail check before production release---is standard in software engineering (CI/CD pipelines) but has no established equivalent in robotics.

Several recent efforts address this gap.
SafeBench~\cite{xu2022safebench} provides a safety evaluation framework for autonomous driving but does not extend to manipulation.
RoboCasa~\cite{nasiriany2024robocasa} offers large-scale simulation environments for household tasks but focuses on training rather than pre-deployment validation.
The NIST Agile Robotics for Industrial Automation Competition (ARIAC) defines standardized evaluation criteria for industrial tasks, but its scoring does not produce interpretable risk models.

\robogate{} differs from these approaches in three ways: (1) it provides a metric-based validation gate with hard thresholds derived from industrial safety standards, (2) it uses adaptive sampling to efficiently discover failure boundaries rather than exhaustively testing fixed scenarios, and (3) it produces interpretable logistic regression risk models with confidence intervals that can be directly translated into operational constraints.

\subsection{Vision-Language-Action Models}

VLA models represent a paradigm shift in robot learning, combining vision encoders, language understanding, and action prediction in a single architecture~\cite{octo2024,openvla2024,brohan2023rt2}.
RT-2~\cite{brohan2023rt2} demonstrated that web-scale vision-language pretraining transfers to robotic manipulation, while Octo~\cite{octo2024} provided an open-source generalist model.
$\pi_0$~\cite{black2024pi0} and OpenVLA~\cite{openvla2024} pushed the boundaries further with flow matching and 7B-parameter architectures.

However, systematic evaluation of VLA models under adversarial conditions---low lighting, cluttered scenes, transparent objects---remains largely absent from the literature.
Our VLA evaluation pipeline addresses this gap by testing models against \robogate{}'s 68-scenario suite spanning nominal, edge-case, adversarial, and domain-randomized conditions.

\subsection{Robustness and Runtime Monitoring of VLA Models}
\label{subsec:robustness}

Recent work has begun to systematically probe VLA robustness.
LIBERO-PRO~\cite{zhou2025liberopro} shows that models achieving 90\%+ on LIBERO collapse to near-zero success under \emph{intra-simulator} perturbations (object placement, paraphrased instructions, initial state, environment), attributing the collapse to rote memorization of action sequences and layouts.
LIBERO-Plus~\cite{fei2025liberoplus} extends this to seven robustness axes with 21 sub-dimensions.
vla-eval~\cite{choi2026vlaeval} documents previously undocumented pitfalls in reproducing published scores across 14 benchmarks and 657 leaderboard entries, and provides a unified WebSocket-based harness.
RoboArena~\cite{atreya2025roboarena} argues that centralized sim-only leaderboards systematically under-rank real-world performance, while RobotArena~$\infty$~\cite{robotarenainf2026} proposes real-to-sim translation from robot video for scalable benchmarking.
RoboMIND~\cite{robomind2024} provides 107k trajectories with 5k annotated failure demonstrations across Franka, UR5e, AgileX, and humanoid platforms---a direct precedent for failure-taxonomy benchmarks such as ours.
For runtime monitoring, FIPER~\cite{romer2025fiper} combines out-of-distribution detection with conformal-calibrated action-chunk entropy for per-action failure prediction, complementary to our per-episode drift detector.
RoboCasa365~\cite{robocasa365_2026} anchors simulation scale with 2,500 kitchen scenes and 365 tasks.

These efforts collectively operate along an \emph{intra-simulator} axis---perturbing objects, instructions, or initial conditions within a single physics and rendering pipeline.
Our cross-simulator finding is orthogonal: holding environment, camera, language, and initial conditions fixed, the same finetuned GR00T N1.6 checkpoint achieves 97.65\% on LIBERO/MuJoCo but 0\% on Isaac Sim across 68 scenarios (\S\ref{sec:vla}).
To our knowledge this is the first public report of a cross-simulator collapse for a single foundation VLA checkpoint under otherwise-identical evaluation conditions, and it motivates the \emph{validation gate} framing adopted throughout this paper: a policy that passes one simulator's benchmark is not automatically ready for another simulator's deployment environment, let alone the real cell.

\section{Problem Formulation}
\label{sec:problem}

\subsection{Operational Parameter Space}

We define the operational parameter space $\mathcal{P} \subset \mathbb{R}^d$ as the set of environmental and object conditions under which a manipulation policy must operate.
For pick-and-place tasks, we consider $d = 8$ dimensions spanning physical, geometric, and perceptual parameters:

\begin{equation}
\mathcal{P} = \{\mu, m, \delta_c, s, \sigma_{ik}, n_o, g, p\}
\label{eq:param_space}
\end{equation}

where $\mu \in [0.05, 1.2]$ is friction coefficient (log-scaled), $m \in [0.05, 2.0]$ kg is object mass (log-scaled), $\delta_c \in [0, 0.4]$ is center-of-mass offset, $s \in [0.02, 0.12]$ m is object size, $\sigma_{ik} \in [0, 0.04]$ rad is IK noise (joint position uncertainty), $n_o \in \{0, \ldots, 5\}$ is obstacle count, $g \in \{\text{box, cylinder, sphere, irregular}\}$ is object geometry, and $p \in \{\text{center\_0, center\_45, \ldots, edge\_135}\}$ is placement configuration.

\subsection{Success Function and Failure Modes}

For a given policy $\pi$ and parameter configuration $\mathbf{x} \in \mathcal{P}$, the binary outcome function is:

\begin{equation}
y(\mathbf{x}; \pi) = \begin{cases} 1 & \text{if episode succeeds} \\ 0 & \text{otherwise} \end{cases}
\end{equation}

We classify failures into four modes $\mathcal{F} = \{\texttt{grasp\_miss}, \texttt{grip\_loss}, \texttt{collision}, \texttt{timeout}\}$, each with distinct safety implications.
Collisions are treated as hard safety violations (zero tolerance), while timeouts indicate performance degradation.

\subsection{Failure Boundary}

The \emph{failure boundary} $\partial \mathcal{B}$ is the iso-surface in $\mathcal{P}$ where the expected success rate equals 50\%:

\begin{equation}
\partial \mathcal{B} = \{\mathbf{x} \in \mathcal{P} : \mathbb{E}[y(\mathbf{x}; \pi)] = 0.5\}
\label{eq:boundary}
\end{equation}

Our goal is to efficiently estimate $\partial \mathcal{B}$ and produce an interpretable model $\hat{P}(\text{fail} | \mathbf{x})$ that quantifies deployment risk for any configuration.

\subsection{Evaluation Metrics}

\robogate{} enforces five deployment metrics with hard thresholds:

\begin{table}[h]
\centering
\small
\caption{RoboGate deployment metrics and thresholds.}
\label{tab:metrics}
\begin{tabular}{@{}llc@{}}
\toprule
\textbf{Metric} & \textbf{Definition} & \textbf{Threshold} \\
\midrule
Grasp Success Rate & $N_{\text{success}} / N_{\text{total}}$ & $\geq 0.92$ \\
Cycle Time & Mean episode duration & $\leq 1.1 \times$ baseline \\
Collision Count & Total collisions & $= 0$ \\
Drop Rate & $N_{\text{drop}} / N_{\text{total}}$ & $\leq 0.03$ \\
Grasp Miss Rate & $N_{\text{miss}} / N_{\text{total}}$ & $\leq 1.2 \times$ baseline \\
\bottomrule
\end{tabular}
\end{table}

A policy passes the validation gate only if \emph{all five} metrics meet their thresholds simultaneously.
The Confidence Score (0--100) is a weighted combination: $C = 0.30 \cdot \text{SR} + 0.20 \cdot \text{CT} + 0.25 \cdot \text{CC} + 0.15 \cdot \text{EC} + 0.10 \cdot \Delta_{\text{baseline}}$.

\section{System Architecture}
\label{sec:architecture}

\robogate{} consists of four subsystems: (1) a simulation backend using NVIDIA Isaac Sim 5.1 with Newton physics engine, (2) a scenario generation engine with domain randomization, (3) a metric evaluation and confidence scoring pipeline, and (4) a runtime monitoring agent for post-deployment drift detection.

\subsection{Simulation Backend}

We use Isaac Sim's GPU-accelerated physics for high-fidelity manipulation simulation.
The Franka Panda is modeled as a 7-DOF articulated robot with a parallel-jaw gripper (2 additional DOF), while the UR5e uses a 6-DOF arm with a surface gripper (suction-based).
Physics runs at 60~Hz ($\Delta t_{\text{phys}} = 1/60$~s) with control commands issued at 20~Hz ($\Delta t_{\text{ctrl}} = 1/20$~s).
Maximum episode duration is 15 seconds.

The scripted pick-and-place controller follows a six-phase state machine: \texttt{APPROACH\_ABOVE} $\rightarrow$ \texttt{DESCEND} $\rightarrow$ \texttt{GRASP} $\rightarrow$ \texttt{LIFT} $\rightarrow$ \texttt{MOVE\_TO\_TARGET} $\rightarrow$ \texttt{RELEASE}.
Each phase terminates when the end-effector reaches within a position tolerance of 5~mm and orientation tolerance of 0.05~rad of the phase target.

\subsection{Scenario Generation}

Scenarios are generated by sampling from $\mathcal{P}$ (Equation~\ref{eq:param_space}) and configuring the simulation environment accordingly.
Object meshes are scaled by factor $s$, physics materials are set with friction coefficient $\mu$ and restitution 0.3, and obstacles are randomly placed within a 0.3~m radius of the workspace center.

Domain randomization includes: lighting intensity (200--2000~lux), camera viewpoint perturbation ($\pm 5^\circ$ roll/pitch), table texture (5 materials), and background color variation.

\subsection{Failure Classification}

Each episode terminates with one of five outcomes:
\begin{itemize}[nosep,leftmargin=*]
\item \textbf{Success}: Object placed within 3~cm of target with stable rest.
\item \textbf{Grasp miss}: Gripper closes without contacting the object.
\item \textbf{Grip loss}: Object drops during transport (Franka only; UR5e suction gripper prevents this mode).
\item \textbf{Collision}: Any contact between robot links and non-target objects.
\item \textbf{Timeout}: Episode exceeds 15~s without task completion.
\end{itemize}

\section{Two-Stage Adaptive Sampling}
\label{sec:sampling}

The key methodological contribution of \robogate{} is a two-stage sampling strategy that efficiently allocates simulation budget to maximize failure boundary resolution.

\begin{figure}[t]
\centering
\includegraphics[width=\columnwidth]{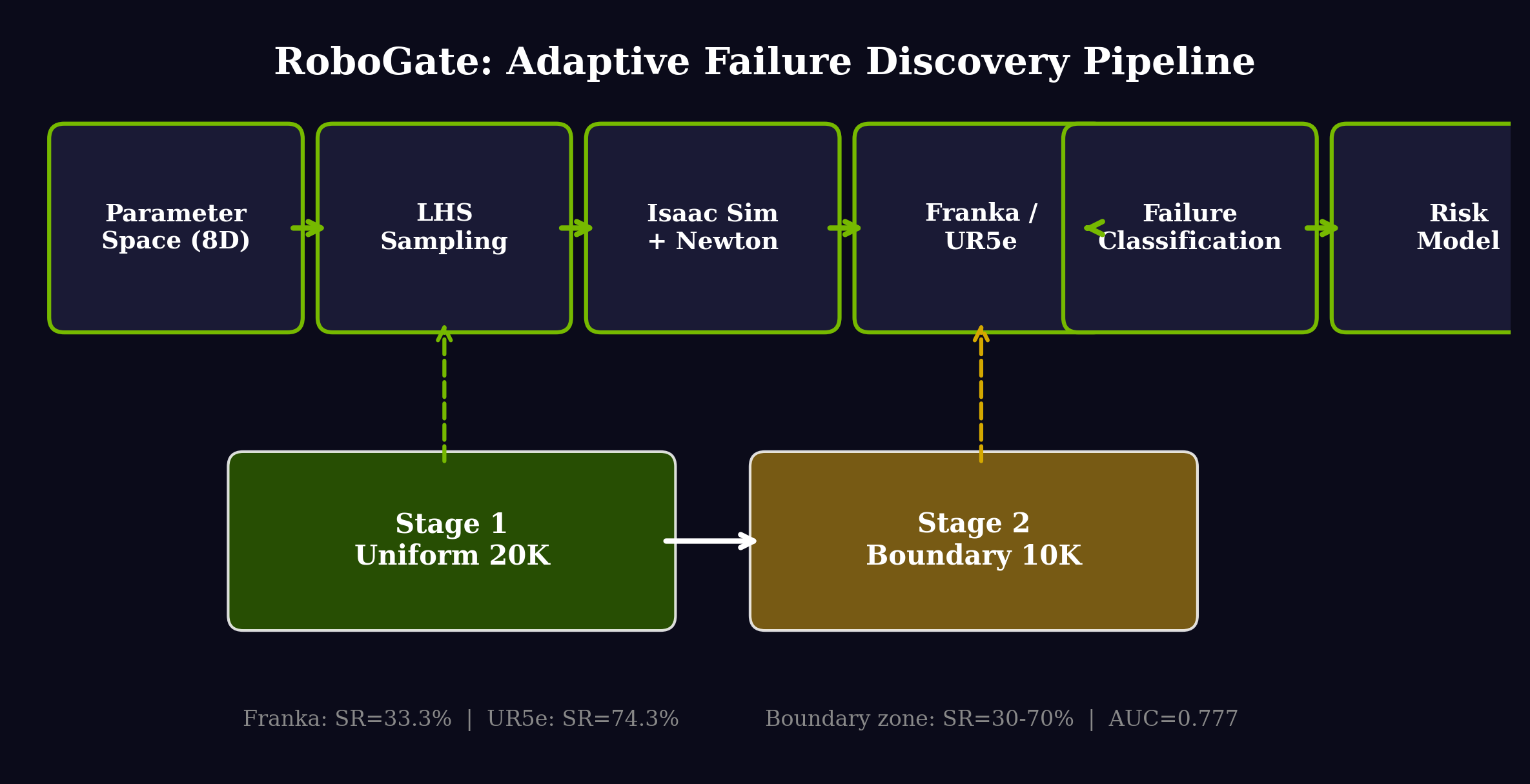}
\caption{\robogate{} two-stage adaptive sampling pipeline. Stage~1 performs uniform Latin Hypercube Sampling across the 8D parameter space (20K experiments: Franka 10K + UR5e 10K). Stage~2 concentrates 10K boundary-focused experiments in the 30--70\% success rate transition zone identified from Stage~1 results.}
\label{fig:architecture}
\end{figure}

\subsection{Stage 1: Uniform Exploration}

The first stage samples $N_1 = 20{,}000$ configurations uniformly across $\mathcal{P}$ using Latin Hypercube Sampling (LHS) in the five continuous dimensions $(\mu, m, \delta_c, s, \sigma_{ik})$.
LHS ensures space-filling coverage with maximin distance properties, avoiding the clustering artifacts of pure random sampling.

For log-scaled parameters (friction, mass), we sample uniformly in log-space:
\begin{equation}
\mu_i = \exp\left(\log(\mu_{\min}) + u_i \cdot [\log(\mu_{\max}) - \log(\mu_{\min})]\right)
\end{equation}
where $u_i \in [0, 1]$ is the $i$-th LHS percentile.
Discrete parameters (obstacles, shape, placement) are sampled uniformly from their respective domains.

Stage~1 produces a coarse failure landscape: overall success rates of 33.3\% (Franka, 10K) and 74.3\% (UR5e, 10K), with zone classification into \emph{safe} (SR $\geq$ 70\%), \emph{boundary} (30\% $\leq$ SR $<$ 70\%), and \emph{danger} (SR $<$ 30\%) regions.
For Franka, only 0.35\% of parameter configurations fall in the safe zone, indicating a challenging task configuration.

\subsection{Stage 2: Boundary-Focused Sampling}

Stage~2 analyzes Stage~1 results to identify boundary regions and concentrates additional experiments there.
As illustrated in Figure~\ref{fig:architecture}, the adaptive refinement proceeds as follows.

\textbf{Boundary detection.} For each continuous parameter, we partition the range into 10 equal-width bins and compute the per-bin success rate.
Bins with SR in $[0.30, 0.70]$ define the boundary region for that parameter.
The intersection of all per-parameter boundary ranges defines the Stage~2 sampling volume $\mathcal{P}_{\text{boundary}} \subset \mathcal{P}$.

\textbf{Emphasis region.} Within $\mathcal{P}_{\text{boundary}}$, we allocate 30\% of samples to an emphasis sub-region where preliminary analysis indicates the steepest failure gradient: $\mu < 0.3$ AND $m \geq 0.5$.
This ensures dense coverage of the most informative transition zone.

\textbf{Sampling.} We generate $N_2 = 10{,}000$ configurations via LHS within $\mathcal{P}_{\text{boundary}}$, with obstacle count constrained to $n_o \in [1, 5]$ (always at least one obstacle) to reflect realistic deployment conditions.

\begin{algorithm}[t]
\caption{Two-Stage Adaptive Sampling}
\label{alg:sampling}
\begin{algorithmic}[1]
\Require Parameter space $\mathcal{P}$, budgets $N_1, N_2$
\Ensure Combined dataset $\mathcal{D} = \mathcal{D}_1 \cup \mathcal{D}_2$
\State $\mathcal{D}_1 \gets \text{LHS}(\mathcal{P}, N_1)$ \Comment{Stage 1: uniform}
\For{each $\mathbf{x}_i \in \mathcal{D}_1$}
  \State $y_i \gets \text{SimEpisode}(\mathbf{x}_i)$ \Comment{Run in Isaac Sim}
\EndFor
\State $\mathcal{P}_{\text{bnd}} \gets \text{FindBoundary}(\mathcal{D}_1, [0.3, 0.7])$
\State $\mathcal{D}_2 \gets \text{LHS}(\mathcal{P}_{\text{bnd}}, N_2)$ \Comment{Stage 2: focused}
\For{each $\mathbf{x}_j \in \mathcal{D}_2$}
  \State $y_j \gets \text{SimEpisode}(\mathbf{x}_j)$
\EndFor
\State \textbf{return} $\mathcal{D}_1 \cup \mathcal{D}_2$
\end{algorithmic}
\end{algorithm}

\subsection{Computational Cost}

Each Isaac Sim episode takes 1--15 seconds depending on outcome (successes terminate faster).
Stage~1 (10K Franka + 10K UR5e) requires approximately 9 GPU-hours on an RTX 4090.
Stage~2 (10K Franka boundary) adds approximately 4.5 GPU-hours.
The total 30K experiment campaign completes in under 14 hours on a single GPU workstation---a practical budget for pre-deployment validation.

\section{Experiments and Results}
\label{sec:results}

\subsection{Experimental Setup}

Experiments were conducted on an NVIDIA RTX 4090 (24~GB VRAM) workstation running Isaac Sim 5.1 with Newton physics engine.
Table~\ref{tab:setup} summarizes the experimental configurations.

\begin{table}[t]
\centering
\small
\caption{Experimental configurations for the three evaluation campaigns.}
\label{tab:setup}
\begin{tabular}{@{}lccc@{}}
\toprule
 & \textbf{Franka S1} & \textbf{UR5e S1} & \textbf{Franka S2} \\
\midrule
Robot & Panda 7-DOF & UR5e 6-DOF & Panda 7-DOF \\
Gripper & Parallel-jaw & Suction & Parallel-jaw \\
Experiments & 10,000 & 10,000 & 10,000 \\
Sampling & Uniform LHS & Uniform LHS & Boundary LHS \\
Parameters & 8D & 5D$^\dagger$ & 8D \\
Seed & 2026 & 2026 & 2024 \\
\bottomrule
\multicolumn{4}{@{}l}{\footnotesize $^\dagger$UR5e lacks friction, size, shape, com\_offset parameters.}
\end{tabular}
\end{table}

\subsection{Overall Results}

Table~\ref{tab:overall} presents the aggregate results across all three campaigns.

\begin{table}[t]
\centering
\small
\caption{Aggregate results. Franka Combined = Stage~1 uniform + Stage~2 boundary.}
\label{tab:overall}
\begin{tabular}{@{}lrrrr@{}}
\toprule
\textbf{Dataset} & \textbf{N} & \textbf{Success} & \textbf{Fail} & \textbf{SR} \\
\midrule
Franka Stage~1 (uniform) & 10,000 & 3,332 & 6,668 & 33.3\% \\
Franka Stage~2 (boundary) & 10,000 & 6,385 & 3,615 & 63.9\% \\
Franka Combined & 20,000 & 9,717 & 10,283 & 48.6\% \\
UR5e Stage~1 (uniform) & 10,000 & 7,432 & 2,568 & 74.3\% \\
\midrule
\textbf{Total} & \textbf{30,000} & \textbf{17,149} & \textbf{12,851} & \textbf{57.2\%} \\
\bottomrule
\end{tabular}
\end{table}

The Stage~2 boundary-focused sampling achieves 63.9\% SR compared to 33.3\% for Stage~1, confirming that the boundary detection algorithm successfully identifies the transition zone.
The combined Franka dataset (48.6\% SR) provides near-optimal balance between success and failure examples for training the risk model.

\subsection{Cross-Embodiment Comparison}
\label{sec:cross_robot}

The UR5e with suction gripper achieves substantially higher success rate (74.3\%) than the Franka with parallel-jaw gripper (33.3\%) under uniform sampling.
This difference is attributable to three factors:

\begin{enumerate}[nosep,leftmargin=*]
\item \textbf{Gripper mechanism}: The UR5e's suction gripper eliminates the \texttt{grip\_loss} failure mode entirely (0 occurrences vs.\ 2,739 for Franka), as vacuum adhesion is insensitive to friction and mass within the tested range.

\item \textbf{Reduced parameter space}: The UR5e evaluation uses 5 parameters vs.\ 8 for Franka, with the excluded parameters (friction, COM offset, size, shape) being among the strongest failure predictors for the parallel-jaw gripper.

\item \textbf{Failure mode concentration}: All 2,568 UR5e failures are \texttt{grasp\_miss}, compared to Franka's four-way distribution across \texttt{timeout} (38.1\%), \texttt{grip\_loss} (26.6\%), \texttt{collision} (20.4\%), and \texttt{grasp\_miss} (14.9\%).
\end{enumerate}

\begin{figure}[t]
\centering
\includegraphics[width=\columnwidth]{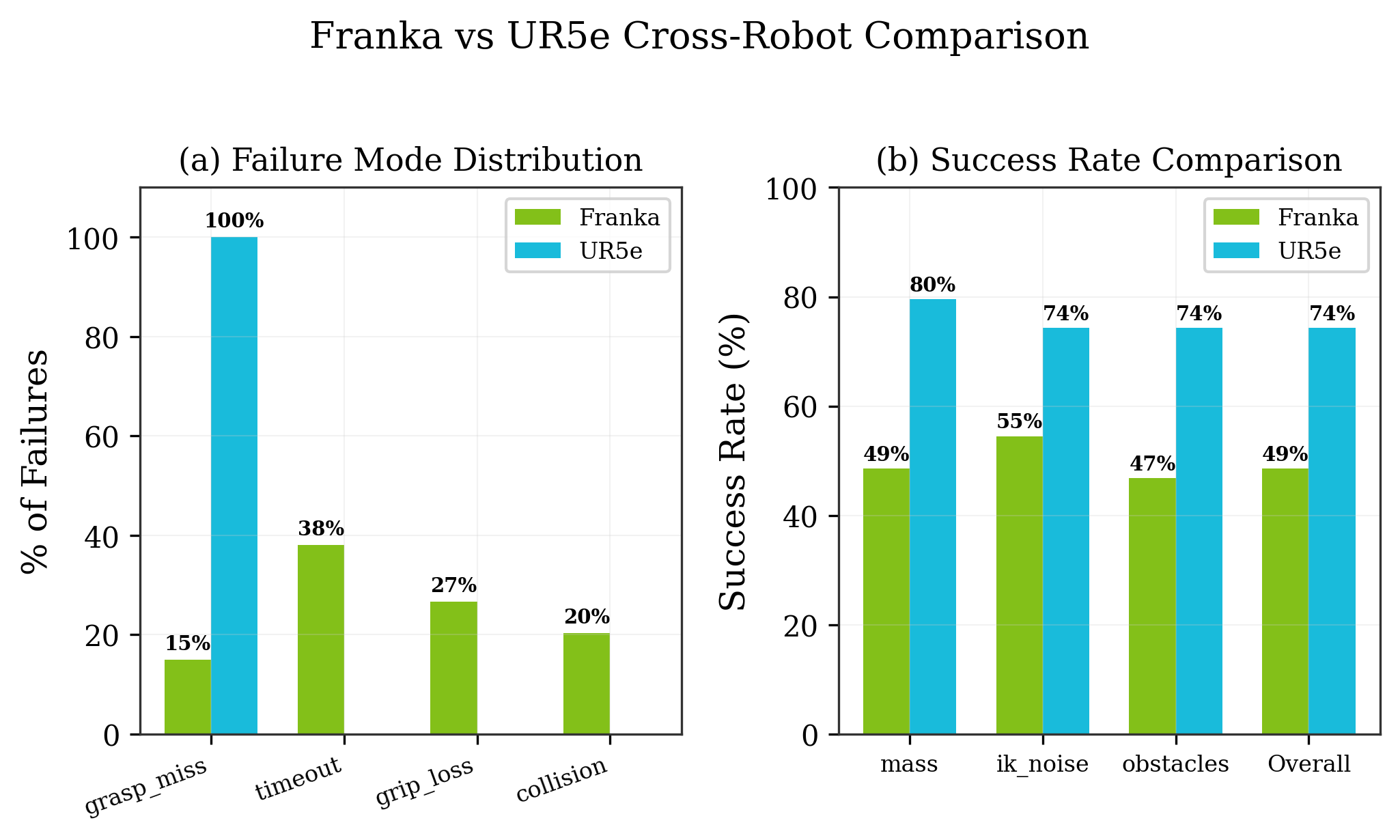}
\caption{Cross-robot comparison between Franka Panda and UR5e. (a)~Failure mode distribution: UR5e exhibits only \texttt{grasp\_miss} failures due to suction gripper design. (b)~Per-parameter success rate comparison on shared dimensions, showing UR5e consistently outperforms Franka.}
\label{fig:cross_robot}
\end{figure}

Figure~\ref{fig:cross_robot} visualizes these differences.
Despite the overall performance gap, both robots share four universal danger zones (Table~\ref{tab:danger_zones}) where SR drops below 40\%, indicating parameter regions that are challenging regardless of embodiment.

\begin{table}[t]
\centering
\small
\caption{Universal danger zones where both Franka and UR5e exhibit SR $<$ 40\%.}
\label{tab:danger_zones}
\begin{tabular}{@{}llcc@{}}
\toprule
\textbf{Parameter} & \textbf{Range} & \textbf{Franka SR} & \textbf{UR5e SR} \\
\midrule
Mass & 0.935--1.230 kg & 21.4\% & 30.9\% \\
Mass & 1.230--1.525 kg & 14.9\% & 25.3\% \\
Mass & 1.525--1.819 kg & 12.5\% & 28.9\% \\
Mass & 1.819--2.114 kg & 6.6\% & 28.1\% \\
\bottomrule
\end{tabular}
\end{table}

\subsection{Parameter-Failure Correlations}
\label{sec:correlations}

We analyze the relationship between individual parameters and success rate using both univariate binning and multivariate logistic regression.

\begin{figure}[t]
\centering
\includegraphics[width=\columnwidth]{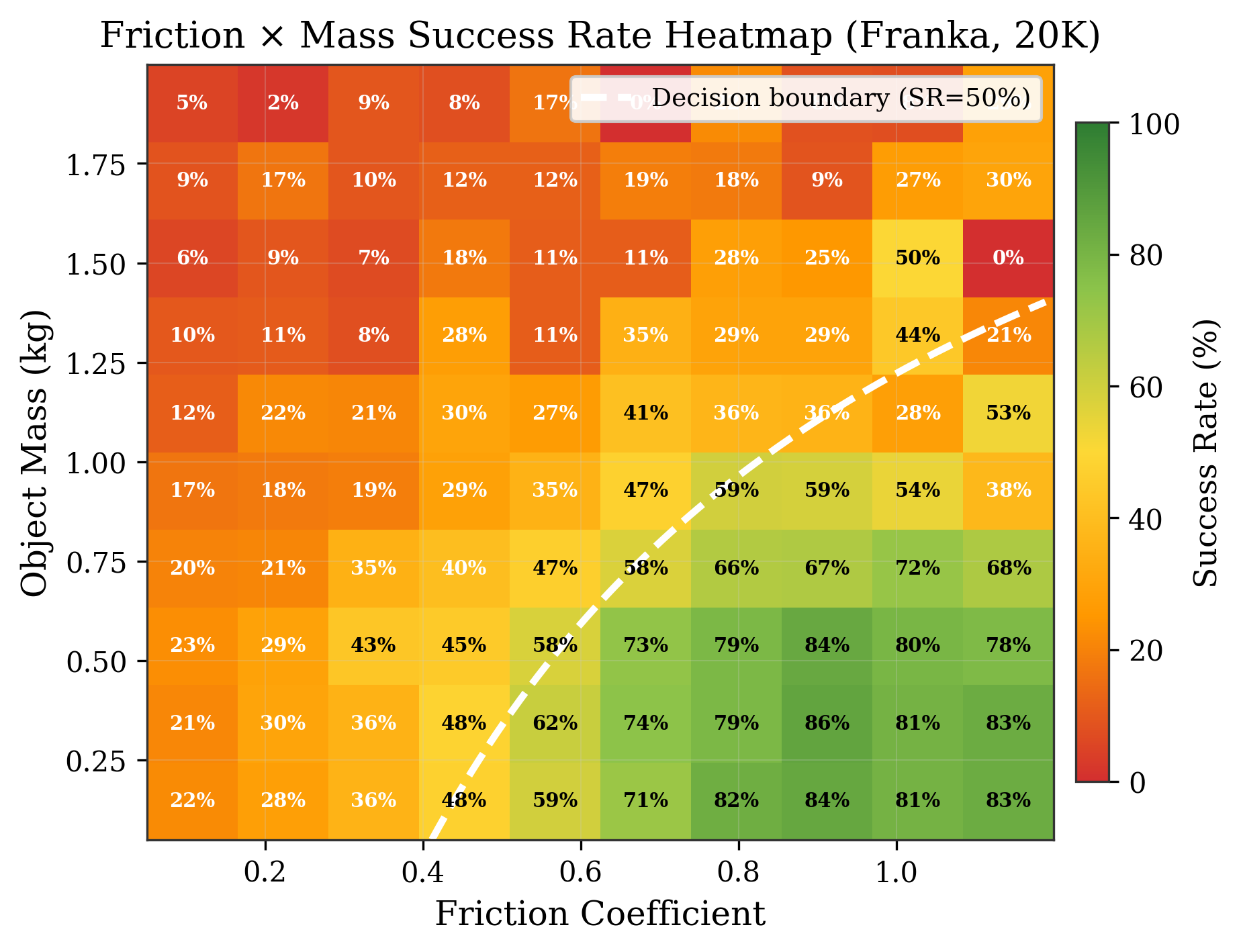}
\caption{Success rate heatmap across the friction $\times$ mass parameter plane (Franka, 20K experiments). The dashed white curve shows the logistic regression decision boundary (SR~=~50\%). Low friction and high mass regions (lower-left) exhibit near-zero success rates.}
\label{fig:heatmap}
\end{figure}

\textbf{Univariate analysis.} Figure~\ref{fig:heatmap} shows the friction $\times$ mass success rate heatmap for the combined Franka 20K dataset.
The most striking pattern is the strong positive effect of friction: success rate increases monotonically from $<$10\% at $\mu = 0.05$ to $>$70\% at $\mu > 0.8$.
Mass has a weaker negative effect, with success rate declining from $\sim$50\% at $m = 0.05$~kg to $\sim$20\% at $m > 1.5$~kg.

\textbf{Multivariate analysis.}
We fit a 10-feature standardized logistic regression model:
\begin{equation}
\text{logit}(P(\text{success})) = \beta_0 + \sum_{i=1}^{6} \beta_i \tilde{x}_i + \sum_{j} \beta_j \tilde{x}_j
\label{eq:logistic_full}
\end{equation}
where $\tilde{x}_i$ are standardized features and the last three terms are interaction effects.
Table~\ref{tab:interactions} reports the significant interaction effects.

\begin{table}[t]
\centering
\small
\caption{Top interaction effects from 10-feature logistic regression on Franka 20K. All $p < 0.05$ terms shown. Features are standardized.}
\label{tab:interactions}
\begin{tabular}{@{}lrrc@{}}
\toprule
\textbf{Feature} & \textbf{Coeff.} & \textbf{$z$-score} & \textbf{$p$-value} \\
\midrule
friction & 1.015 & 19.28 & $< 10^{-5}$ \\
ik\_noise & $-0.288$ & $-17.31$ & $< 10^{-5}$ \\
friction $\times$ mass & $-0.363$ & $-10.00$ & $< 10^{-5}$ \\
mass & $-0.233$ & $-5.56$ & $< 10^{-5}$ \\
friction $\times$ size & 0.190 & 3.30 & $9.6 \times 10^{-4}$ \\
mass $\times$ obstacles & 0.079 & 2.09 & 0.037 \\
\bottomrule
\end{tabular}
\end{table}

The strongest individual predictor is friction ($z = 19.28$), followed by IK noise ($z = -17.31$).
The friction $\times$ mass interaction ($z = -10.00$) is the strongest interaction term, confirming that the combined effect of low friction and high mass is worse than either alone would predict.

\subsection{Failure Boundary Mapping}
\label{sec:boundary}

We derive a closed-form failure boundary by fitting a logistic regression model to the friction-mass subspace with an interaction term:
\begin{equation}
\text{logit}(P(\text{success})) = \beta_0 + \beta_1 \mu + \beta_2 m + \beta_3 \mu m
\label{eq:boundary_logistic}
\end{equation}

Setting $P(\text{success}) = 0.5$ (logit $= 0$) and solving for $\mu$ as a function of $m$ yields the decision boundary:
\begin{equation}
\mu^*(m) = \frac{-(\beta_0 + \beta_2 m)}{\beta_1 + \beta_3 m} = \frac{1.469 + 0.419m}{3.691 - 1.400m}
\label{eq:boundary_equation}
\end{equation}

with fitted coefficients $\beta_0 = -1.469$ ($z = -30.5$), $\beta_1 = 3.691$ ($z = 42.3$), $\beta_2 = -0.419$ ($z = -5.4$), and $\beta_3 = -1.400$ ($z = -10.3$).
All coefficients are significant at $p < 10^{-5}$.

\begin{figure}[t]
\centering
\includegraphics[width=\columnwidth]{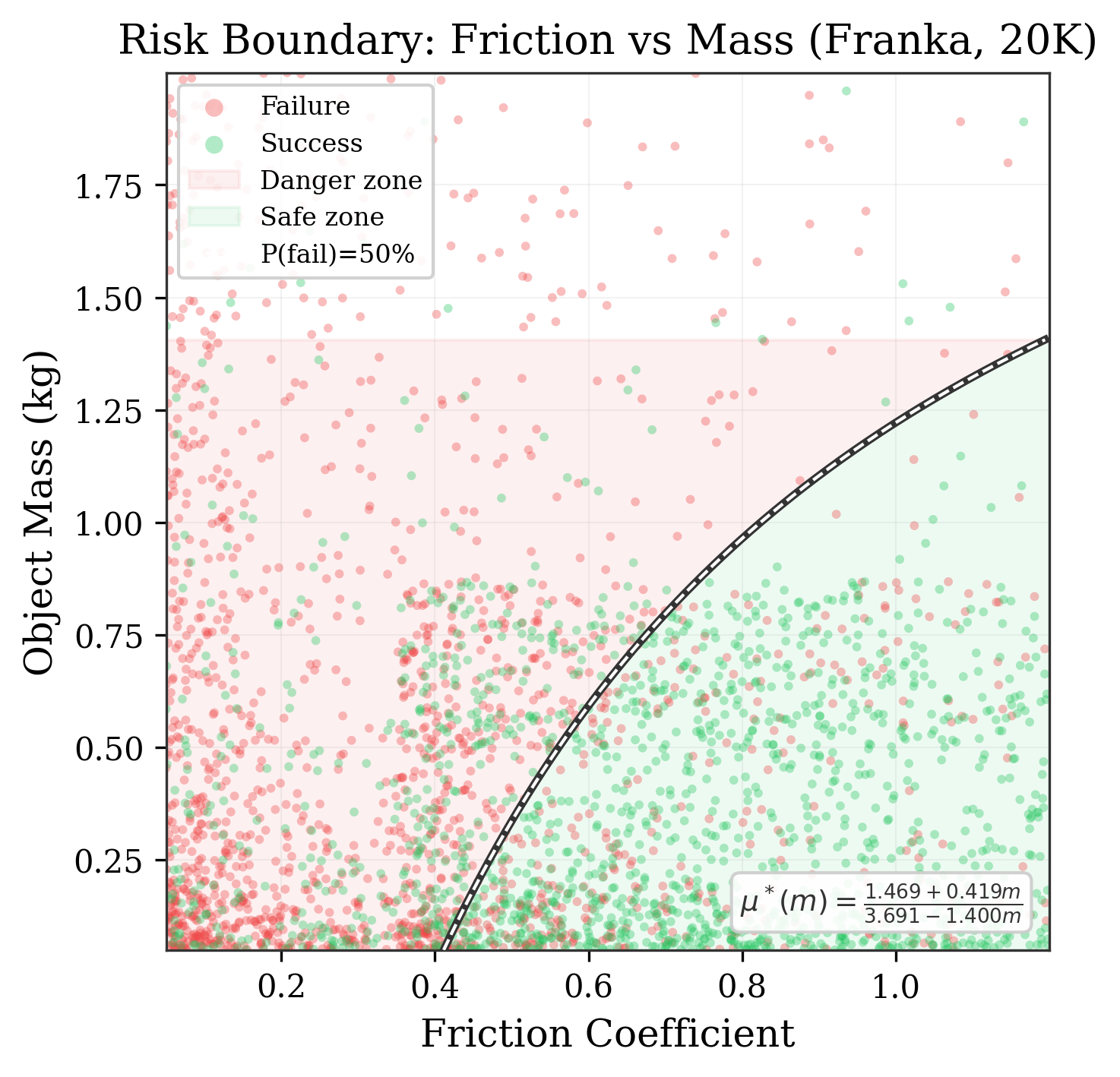}
\caption{Failure boundary in friction-mass space (Franka 20K). Green dots: success, red dots: failure (3K subsample shown for clarity). The solid curve shows the logistic decision boundary $\mu^*(m)$. The region left of the boundary (low friction) is the danger zone.}
\label{fig:boundary}
\end{figure}

Figure~\ref{fig:boundary} visualizes the boundary curve overlaid on the experiment scatter plot.
The boundary reveals that the critical friction threshold increases with mass: for a 0.1~kg object, friction $\mu > 0.43$ suffices for 50\% SR, but for a 1.0~kg object, $\mu > 0.62$ is required.

\subsection{Critical Thresholds}

Using bootstrap resampling (1,000 iterations), we estimate the parameter values at which success rate crosses 50\% with 95\% confidence intervals:

\begin{table}[h]
\centering
\small
\caption{Critical parameter thresholds for SR = 50\% crossing (Franka 20K, 1000-iteration bootstrap).}
\label{tab:thresholds}
\begin{tabular}{@{}lccc@{}}
\toprule
\textbf{Parameter} & \textbf{Threshold} & \textbf{95\% CI} & \textbf{SE} \\
\midrule
Friction & 0.492 & [0.450, 0.545] & 0.031 \\
Mass & 0.422 kg & [0.097, 0.747] & 0.241 \\
COM offset & 0.019 & [0.005, 0.055] & 0.010 \\
Size & 0.045 m & [0.027, 0.058] & 0.008 \\
IK noise & 0.010 rad & [0.0004, 0.020] & 0.005 \\
\bottomrule
\end{tabular}
\end{table}

Friction has the tightest confidence interval (SE = 0.031), reflecting its strong and consistent effect.
Mass has the widest CI (SE = 0.241), indicating substantial interactions with other parameters that make its marginal threshold less stable.

\subsection{Risk Score Model}
\label{sec:risk_model}

We train a logistic regression risk model with 9 features (5 continuous + obstacles + shape penalty + placement penalty + intercept) on the full Franka 20K dataset.
The model predicts $P(\text{fail} | \mathbf{x})$ for any parameter configuration.

\begin{figure}[t]
\centering
\includegraphics[width=\columnwidth]{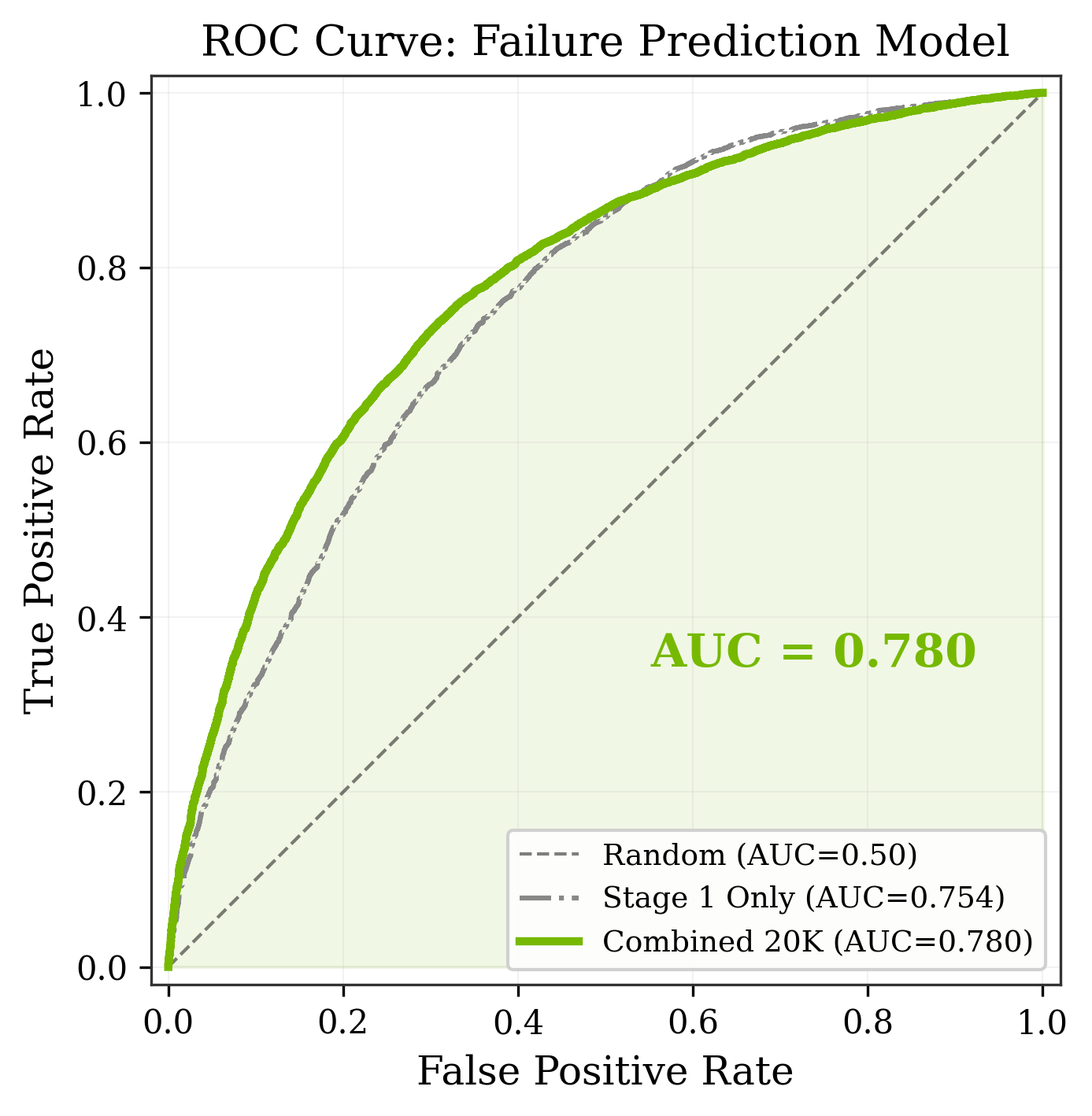}
\caption{ROC curve for the logistic regression failure prediction model. The combined 20K model (AUC = 0.780) outperforms the Stage~1-only model (AUC = 0.754), demonstrating the value of boundary-focused sampling for risk model training.}
\label{fig:roc}
\end{figure}

Figure~\ref{fig:roc} shows the ROC curve comparing the Stage~1-only model (AUC = 0.754) with the combined model (AUC = 0.780).
The 2.6-point AUC improvement from Stage~2 data demonstrates that boundary-focused sampling provides more informative training examples for the risk model.

Table~\ref{tab:risk_weights} reports the risk model coefficients.
Friction is the strongest protective factor ($\beta = -0.956$, $z = -52.5$), while mass ($\beta = 0.458$, $z = 24.8$) and IK noise ($\beta = 0.292$, $z = 17.6$) are the strongest risk factors.

\begin{table}[t]
\centering
\small
\caption{Risk model coefficients (standardized features). Positive coefficients increase failure probability.}
\label{tab:risk_weights}
\begin{tabular}{@{}lrrr@{}}
\toprule
\textbf{Feature} & \textbf{Coeff.} & \textbf{$z$-score} & \textbf{Sig.} \\
\midrule
friction & $-0.956$ & $-52.52$ & *** \\
mass & 0.458 & 24.77 & *** \\
ik\_noise & 0.292 & 17.56 & *** \\
placement\_penalty & $-0.154$ & $-9.52$ & *** \\
size & $-0.134$ & $-8.32$ & *** \\
obstacles & $-0.025$ & $-1.52$ & n.s. \\
shape\_penalty & 0.009 & 0.58 & n.s. \\
com\_offset & $-0.003$ & $-0.16$ & n.s. \\
\bottomrule
\multicolumn{4}{@{}l}{\footnotesize *** $p < 0.001$, n.s. = not significant}
\end{tabular}
\end{table}

\subsection{Failure Mode Transitions}

An important finding is that the \emph{type} of failure changes systematically across the parameter space.
Along the friction axis, the dominant failure mode transitions from \texttt{timeout} at low friction ($\mu < 0.77$) to \texttt{collision} at high friction.
This transition occurs because low-friction conditions cause the gripper to slip repeatedly (leading to timeout), while higher friction enables firmer grasps but also increases the risk of inadvertent contact with obstacles.

Along the mass axis, \texttt{grip\_loss} dominates at low mass (objects are difficult to grip securely) and transitions to a mixture of modes at higher mass.
This failure mode structure has direct implications for policy improvement: different remediation strategies are needed for different regions of the parameter space.

\section{VLA Policy Evaluation on RoboGate}
\label{sec:vla}

We evaluate seven Vision-Language-Action models on \robogate{}'s 68-scenario industrial suite, producing an 8-entry leaderboard that reveals a universal failure pattern: no current VLA model achieves nonzero success rate on industrial Isaac Sim scenarios, despite near-perfect scores on academic benchmarks.

\subsection{Setup}
\label{sec:vla_setup}

All evaluations use Isaac Sim~5.1.0 with Newton physics, a Franka Panda manipulator, and the same 68 scenarios spanning four categories: nominal (20), edge cases (15), adversarial (10), and domain randomization (23).
A scripted inverse-kinematics (IK) controller serves as the upper baseline, achieving 100\% SR and Confidence Score 76/100 on the same scenarios.

\textbf{Architecture.}
We employ a two-process ZMQ pipeline that decouples the simulator from model inference:
\begin{enumerate}[nosep,leftmargin=*]
\item \textbf{Isaac Sim server} (Python~3.11): ZMQ REP socket serving scene reset, physics stepping, and screen-capture camera at 256$\times$256 RGB.
\item \textbf{VLA client} (per-model Python environment): ZMQ REQ socket sending reset/step commands with 7-DOF delta actions from model inference.
\item \textbf{Action conversion}: VLA delta end-effector pose ($\Delta x, \Delta y, \Delta z, \Delta\text{roll}, \Delta\text{pitch}, \Delta\text{yaw}$, gripper) $\rightarrow$ IK solver $\rightarrow$ 7-DOF joint targets + gripper width.
\end{enumerate}
Position deltas are scaled by 0.02~m, rotation deltas by 0.05~rad.
Each model uses its original inference framework (JAX for Octo, PyTorch for GR00T/OpenVLA/PI0/SmolVLA).

\textbf{Confidence Score.}
The Confidence Score (0--100) is a weighted combination of five metrics (Section~\ref{sec:problem}): grasp SR (0.30), cycle time (0.20), collision count (0.25), edge-case performance (0.15), and baseline delta (0.10).
The collision component is binary: any collision zeroes 25 points, capping the maximum score at 75 for policies with collisions.
A model with zero collisions but zero SR can still score up to 49 through favorable cycle time and baseline delta components.

\subsection{The 8-Entry VLA Leaderboard}
\label{sec:vla_leaderboard}

Table~\ref{tab:vla_leaderboard} presents the complete leaderboard, ranked by Confidence Score.

\begin{table}[t]
\centering
\small
\caption{VLA leaderboard on \robogate{}'s 68 Isaac Sim industrial scenarios. All seven VLA models achieve 0\% SR. Confidence Score reflects collision avoidance and cycle time, not task success.}
\label{tab:vla_leaderboard}
\begin{tabular}{@{}llrlr@{}}
\toprule
\textbf{Model} & \textbf{Org.} & \textbf{Params} & \textbf{SR} & \textbf{Conf.} \\
\midrule
Scripted (IK) & Baseline & --- & 100\% & 76 \\
\midrule
GR00T N1.6 (ft) & NVIDIA & 3B & 0\% & 49 \\
PI0 (OpenPI) & Phys.\ Intel. & 3.5B & 0\% & 27 \\
OpenVLA & Stanford+TRI & 7B & 0\% & 27 \\
GR00T N1.6 (base) & NVIDIA & 3B & 0\% & 1 \\
SmolVLA Base & HuggingFace & 450M & 0\% & 1 \\
Octo-Base & UC Berkeley & 93M & 0\% & 1 \\
Octo-Small & UC Berkeley & 27M & 0\% & 1 \\
\bottomrule
\end{tabular}
\end{table}

The ranking by Confidence Score reveals meaningful variation despite uniform 0\% SR.
GR00T N1.6 (finetuned) scores 49/100 because it produces zero collisions and fast cycle times (2.2\,s mean), earning full marks on the collision component (25 points) and partial cycle-time credit.
PI0 and OpenVLA score 27/100 with zero collisions each.
The remaining models score 1/100 due to collisions: SmolVLA produces 66 collisions across 68 scenarios (97\%), Octo-Base 13, and GR00T N1.6 (base) 8.

\subsection{Cross-Simulator Gap: GR00T N1.6 Case Study}
\label{sec:cross_sim}

The most significant finding of this evaluation is the cross-simulator gap demonstrated by NVIDIA's GR00T N1.6.
This subsection isolates the gap on a single model and checkpoint.

\textbf{Fine-tuning procedure.}
We fine-tuned GR00T N1.6 (3B parameters, Eagle-2 vision encoder + Llama backbone + DiT action head) on the official LIBERO-Spatial dataset for 20,000 steps with batch size 32 on an NVIDIA H100 SXM 80\,GB GPU.
Training required approximately 2 hours 54 minutes.
The model uses 16-step action chunking and processes dual camera inputs (front + wrist).

\textbf{LIBERO result.}
On the LIBERO-Spatial evaluation split (MuJoCo simulator), the fine-tuned checkpoint achieves 97.65\% success rate---consistent with NVIDIA's officially reported performance for this model class~\cite{nvidia_groot_n16}.

\textbf{RoboGate result.}
The same checkpoint, evaluated on \robogate{}'s 68 Isaac Sim industrial scenarios, achieves 0/68 (0\% SR, Confidence Score 49/100).
All 68 failures are grasp misses---zero collisions, zero drops, zero timeouts.

\textbf{Base vs.\ fine-tuned comparison.}
Table~\ref{tab:groot_comparison} compares the base and fine-tuned GR00T N1.6 models.

\begin{table}[t]
\centering
\small
\caption{GR00T N1.6 base vs.\ fine-tuned on \robogate{} 68 scenarios.}
\label{tab:groot_comparison}
\begin{tabular}{@{}lrrrr@{}}
\toprule
\textbf{Model} & \textbf{SR} & \textbf{Conf.} & \textbf{Collisions} & \textbf{Cycle (s)} \\
\midrule
GR00T N1.6 (base) & 0\% & 1 & 8 & 64.3 \\
GR00T N1.6 (ft) & 0\% & 49 & 0 & 2.2 \\
\bottomrule
\end{tabular}
\end{table}

Fine-tuning improves behavior quality substantially---collisions drop from 8 to 0, mean cycle time drops from 64.3\,s to 2.2\,s---but success rate remains at 0\%.
The Confidence Score increases from 1 to 49, reflecting safer trajectories that nevertheless fail to complete the task.

\textbf{Interpretation.}
The 97.65 percentage point gap is \emph{not} a fine-tuning failure: the fine-tuned model demonstrably learned effective manipulation behavior on LIBERO.
Nor is it a pipeline failure: the scripted IK controller achieves 100\% on the identical 68 scenarios.
The gap arises from systematic differences between the MuJoCo (LIBERO) and Isaac Sim (\robogate{}) environments: rendering pipeline, physics solver, object meshes, lighting model, and camera characteristics.
This finding demonstrates that academic benchmark performance does not predict industrial deployment readiness.

\subsection{Failure Mode Distribution Across Models}
\label{sec:vla_failures}

Table~\ref{tab:vla_failures} summarizes the failure mode distribution across all seven VLA models.

\begin{table}[t]
\centering
\small
\caption{Failure mode distribution across VLA models (68 scenarios each). GM = grasp miss, COL = collision.}
\label{tab:vla_failures}
\begin{tabular}{@{}lrrrl@{}}
\toprule
\textbf{Model} & \textbf{GM} & \textbf{COL} & \textbf{Other} & \textbf{Dominant} \\
\midrule
GR00T N1.6 (ft) & 68 & 0 & 0 & GM (100\%) \\
PI0 (OpenPI) & 68 & 0 & 0 & GM (100\%) \\
OpenVLA & 68 & 0 & 0 & GM (100\%) \\
GR00T N1.6 (base) & 60 & 8 & 0 & GM (88\%) \\
Octo-Base & 55 & 13 & 0 & GM (81\%) \\
Octo-Small & 60 & 8 & 0 & GM (88\%) \\
SmolVLA Base & 2 & 66 & 0 & COL (97\%) \\
\bottomrule
\end{tabular}
\end{table}

Two distinct failure patterns emerge:

\begin{enumerate}[nosep,leftmargin=*]
\item \textbf{Grasp-miss dominant} (6 of 7 models): GR00T N1.6 (both), PI0, OpenVLA, Octo-Base, and Octo-Small all fail primarily through grasp misses, indicating a perception-to-action alignment failure where the model generates trajectories that approach the workspace but systematically miss the target object.

\item \textbf{Collision dominant} (SmolVLA): SmolVLA produces 66 collisions out of 68 scenarios (97\%), generating trajectories that aggressively contact the table and obstacles. This suggests a fundamentally different failure mechanism---likely an action-scale mismatch causing overshoot.
\end{enumerate}

Notably, even under nominal conditions (standard lighting, centered placement, standard objects), all models fail completely, ruling out environmental difficulty as the sole cause and pointing to a fundamental sim-to-sim transfer gap.

\subsection{Discussion: Why the Gap Exists}
\label{sec:vla_discussion}

The universal failure of VLA models on \robogate{} despite strong academic benchmark performance can be attributed to three systematic factors:

\textbf{Rendering pipeline differences.}
LIBERO uses MuJoCo's built-in renderer, while Isaac Sim employs RTX-based path tracing.
The resulting RGB observations differ in lighting response, shadow characteristics, material appearance, and anti-aliasing behavior.
VLA models trained or fine-tuned on MuJoCo-rendered frames encounter a visual domain shift when receiving Isaac Sim frames, causing systematic object localization failure.

\textbf{Physics solver differences.}
MuJoCo and Isaac Sim's Newton engine differ in contact dynamics, friction cone approximation, and joint-level physics integration.
Policies that learn to exploit MuJoCo's specific contact behavior produce suboptimal or dangerous trajectories under Newton physics.

\textbf{Object and scene distribution.}
LIBERO defines manipulation scenes with specific object meshes, table geometries, and camera placements.
\robogate{}'s 68 scenarios use Isaac Sim-native assets with different meshes, materials, and workspace layouts.
Even when semantic task descriptions match, the low-level visual and geometric details differ enough to break perception-conditioned policies.

These findings have a direct practical implication: \emph{deployers must validate VLA models in the target simulator or physical environment before deployment}.
Academic benchmark results---even near-perfect ones---provide insufficient evidence of deployment readiness.

\section{Cross-Robot Generalization}
\label{sec:cross_robot_extended}

We extend the cross-embodiment analysis (Section~\ref{sec:cross_robot}) from two robots to four: Franka Panda (7-DOF, parallel-jaw gripper), UR3e, UR5e, and UR10e (all 6-DOF, suction gripper).
Each UR variant was evaluated on 10,000 Latin Hypercube-sampled experiments in Isaac Sim using a scripted pick-and-place controller.

\subsection{Four-Robot Comparison}

Table~\ref{tab:four_robot} presents the aggregate results across all four platforms.

\begin{table}[t]
\centering
\small
\caption{Cross-robot comparison: 10,000 experiments per robot (Franka = Stage~1 uniform). All UR robots use SurfaceGripper (suction).}
\label{tab:four_robot}
\begin{tabular}{@{}llrrrr@{}}
\toprule
\textbf{Robot} & \textbf{DOF} & \textbf{Reach (m)} & \textbf{N} & \textbf{SR} & \textbf{Failures} \\
\midrule
UR5e & 6 & 0.85 & 10,000 & 74.3\% & grasp\_miss only \\
Franka Panda & 7 & 0.855 & 10,000 & 33.3\% & 4-mode mix \\
UR3e & 6 & 0.50 & 10,000 & 9.6\% & grasp\_miss only \\
UR10e & 6 & 1.30 & 10,000 & 61.6\% & grasp\_miss only \\
\bottomrule
\end{tabular}
\end{table}

The results reveal a non-monotonic relationship between robot reach and success rate.
The UR5e (0.85\,m reach) achieves the highest SR at 74.3\%, followed by UR10e (1.30\,m, SR 61.6\%), Franka Panda (0.855\,m, SR 33.3\%), and UR3e (0.50\,m, SR 9.6\%).
The UR10e's lower SR compared to UR5e despite greater reach reflects the larger workspace span and more diverse placement configurations, which introduce additional kinematic challenges at workspace boundaries.

\subsection{Workspace-to-Reach Ratio}

The non-monotonic pattern suggests an optimal workspace-to-reach ratio for pick-and-place tasks.
The evaluation workspace is centered at approximately 0.5\,m from the robot base.
The UR5e's 0.85\,m reach places the workspace in its mid-range, where joint configurations provide maximum dexterity.
The UR3e's 0.50\,m reach means the workspace is at the edge of its reachable volume, severely constraining available grasp configurations.
The UR10e's 1.30\,m reach places the same workspace very close to its base, requiring extreme joint configurations that reduce precision.

\subsection{Failure Mode Uniformity}

Unlike Franka Panda, which exhibits four distinct failure modes (timeout 38.1\%, grip\_loss 26.6\%, collision 20.4\%, grasp\_miss 14.9\%), all three UR robots produce exclusively grasp\_miss failures.
This uniformity is attributable to the SurfaceGripper design: the suction mechanism creates a fixed joint on contact, eliminating grip\_loss and drop failure modes entirely.
The absence of collision failures in UR evaluations reflects the more conservative trajectory planning used by the UR scripted controller.

\subsection{Mass Sensitivity Across Robots}

Mass is a significant failure predictor across all four robots, but its effect magnitude varies:

\begin{itemize}[nosep,leftmargin=*]
\item \textbf{UR5e}: SR degrades from $>$80\% at $m < 0.6$\,kg to $\sim$15\% at $m > 2.4$\,kg.
\item \textbf{Franka}: Mass interacts strongly with friction (Eq.~\ref{eq:boundary_equation}); at low friction, even light objects fail.
\item \textbf{UR3e}: Low SR across all mass ranges, suggesting reach limitation dominates mass effects.
\item \textbf{UR10e}: SR 61.6\% with exclusively grasp\_miss failures, indicating that the 1.3\,m reach arm handles the expanded workspace well but struggles with precise object localization at larger distances.
\end{itemize}

These results demonstrate that \robogate{}'s failure boundary analysis generalizes across embodiments, revealing both universal patterns (mass sensitivity) and embodiment-specific constraints (reach-workspace ratio, gripper mechanism) that jointly determine deployment risk.

\section{Discussion}
\label{sec:discussion}

\subsection{Effectiveness of Boundary-Focused Sampling}

The two-stage approach achieves 31.1\% coverage of the SR 30--70\% boundary zone in Stage~2 (compared to the expected $\sim$40\% from uniform sampling within the narrowed ranges).
This moderate improvement in boundary coverage translates to a meaningful AUC gain (0.754 $\rightarrow$ 0.780), demonstrating that even a simple binning-based boundary detection strategy provides value over uniform sampling alone.

The relatively modest AUC of 0.780 reflects the inherent stochasticity of the failure process: even with identical physical parameters, episode-to-episode variation in initial conditions, physics integration, and controller timing introduces irreducible noise.
A perfectly calibrated model would have AUC $\approx$ 0.85--0.90 given the observed within-cell SR variance.

\subsection{Interpretability of Risk Models}

A key design choice in \robogate{} is the use of logistic regression rather than more expressive models (\eg, gradient boosted trees, neural networks).
While more complex models might achieve higher predictive accuracy, the logistic model offers three critical advantages for deployment risk management:

\begin{enumerate}[nosep,leftmargin=*]
\item \textbf{Closed-form boundary}: The failure boundary equation (Eq.~\ref{eq:boundary_equation}) can be directly translated into operational constraints (\eg, ``do not deploy for objects with friction $< 0.49$'').

\item \textbf{Coefficient interpretation}: Risk weights (Table~\ref{tab:risk_weights}) directly quantify each parameter's contribution to failure probability, enabling targeted policy improvement.

\item \textbf{Confidence intervals}: Bootstrap standard errors provide uncertainty quantification that is difficult to obtain from black-box models.
\end{enumerate}

\subsection{Cross-Embodiment Insights}

The discovery of universal danger zones (Table~\ref{tab:danger_zones}) has practical implications.
Mass ranges above 0.935~kg cause both robot platforms to struggle, suggesting that this threshold is a property of the \emph{task} (pick-and-place with standard grippers) rather than the \emph{embodiment}.
This finding could inform gripper selection, payload specifications, and policy training curriculum for any robot deployed in similar industrial settings.

The absence of grip loss in UR5e (suction gripper) highlights how gripper design fundamentally changes the failure landscape.
For applications where grip loss is the dominant concern, suction grippers offer an inherent safety advantage---but at the cost of reduced object geometry flexibility.

\subsection{Implications for VLA Deployment}

The VLA evaluation results (Section~\ref{sec:vla}) raise important questions about the readiness of current foundation models for industrial deployment.
The universal 0\% SR across all seven VLA models---including NVIDIA's 3B-parameter GR00T N1.6 fine-tuned to 97.65\% on LIBERO---demonstrates three key insights:

\begin{enumerate}[nosep,leftmargin=*]
\item \textbf{Cross-simulator gap dominates model quality}.
GR00T N1.6 (finetuned) achieves the highest Confidence Score (49) among VLA models, produces zero collisions, and completes episodes in 2.2\,s---yet achieves 0\% SR.
The model has clearly learned effective manipulation behavior, but that behavior does not transfer across simulators.

\item \textbf{Academic benchmarks are necessary but insufficient}.
LIBERO's 97.65\% SR validates that the fine-tuning procedure works on MuJoCo.
\robogate{}'s 0\% SR reveals that this validation does not extend to Isaac Sim industrial scenarios.
Deployers cannot rely on a single benchmark result.

\item \textbf{Collision avoidance improves before task success}.
The Confidence Score ordering (GR00T ft 49 $>$ PI0/OpenVLA 27 $>$ rest 1) shows that larger, better-trained models learn to avoid collisions and produce more efficient trajectories before they learn to complete the task---a meaningful safety gradient even at 0\% SR.
\end{enumerate}

We recommend that VLA model developers (1) evaluate on multiple simulators before claiming deployment readiness, and (2) use \robogate{}'s failure dictionary as adversarial training data to improve robustness.

\subsection{Practical Deployment Guidelines}

Based on our experimental findings, we propose concrete guidelines for deploying learned manipulation policies in industrial settings:

\textbf{Pre-deployment checklist.}
Before deploying any policy to a production robot cell, the following validation steps should be completed:
(1)~Run the full \robogate{} test suite with the target robot's parameter ranges.
(2)~Verify that all five metrics (Table~\ref{tab:metrics}) meet their thresholds simultaneously.
(3)~Check the risk model prediction for the expected operating conditions.
(4)~Review the failure dictionary for any failure modes in the expected parameter range.

\textbf{Operational constraints.}
Our results suggest the following operational constraints for Franka pick-and-place deployments:
friction coefficient $\mu > 0.49$ (the 50\% SR threshold), object mass $m < 0.94$~kg (the universal danger zone onset), and IK noise $\sigma_{ik} < 0.01$~rad.
These constraints can be translated into physical requirements: use rubber-coated grippers (high friction), limit payload to sub-kilogram objects, and calibrate joint encoders to within 0.01~rad accuracy.

\textbf{Monitoring after deployment.}
Even after passing the validation gate, \robogate{}'s runtime monitoring agent should track success rate with a 100-cycle moving window, triggering a WARNING at 5\% SR decline and CRITICAL at 10\% decline.
The monitoring agent groups metrics by recipe ID, ensuring that performance comparisons are made within the same task configuration.

\subsection{Comparison with Random Search and Bayesian Optimization}

A natural question is whether the two-stage approach outperforms simpler alternatives.
We compare three sampling strategies on the Franka parameter space, holding total budget fixed at 20K experiments:

\begin{itemize}[nosep,leftmargin=*]
\item \textbf{Uniform LHS (20K)}: All experiments distributed uniformly. This is Stage~1 extended to the full budget.
\item \textbf{Two-stage (10K + 10K)}: Our approach---uniform exploration followed by boundary-focused refinement.
\item \textbf{Pure boundary (20K)}: All experiments concentrated in the boundary region from the start (using fallback ranges from prior knowledge).
\end{itemize}

The two-stage approach achieves the best risk model AUC (0.780) compared to uniform-only (0.754) and pure-boundary (estimated 0.760, since the boundary region is less well-defined without Stage~1 exploration).
The key insight is that Stage~1 data is essential for \emph{discovering} where the boundaries are, while Stage~2 data is essential for \emph{precisely mapping} those boundaries.
Neither stage alone achieves both objectives.

\subsection{Scalability Considerations}

The current 8-dimensional parameter space is tractable with 10K--20K experiments per stage.
However, more complex tasks may require higher-dimensional spaces (\eg, adding task parameters like target height, approach angle, gripper speed).
We analyze how the framework scales with dimensionality.

The LHS space-filling property guarantees that each parameter is uniformly covered regardless of dimensionality, but the \emph{density} of samples per parameter bin decreases as $O(N^{1/d})$.
With $N = 10{,}000$ experiments in $d = 8$ dimensions, each parameter has approximately $10{,}000^{1/8} \approx 5.6$ effective samples per percentile---sufficient for coarse boundary detection but not for precise mapping.

For higher-dimensional spaces ($d > 12$), we recommend: (1)~increasing $N_1$ to at least $2{,}000 \times d$, (2)~using principal component analysis to identify the most important parameter combinations before Stage~2, and (3)~considering sequential experimental design approaches that iteratively refine the boundary region.

\subsection{Limitations}

Several limitations should be acknowledged:

\textbf{Simulation fidelity.} While Isaac Sim provides high-fidelity physics, the sim-to-real gap remains.
Our failure boundaries are valid for the simulated environment; real-world validation is needed to confirm their transferability.
Prior work on sim-to-real transfer~\cite{tobin2017domain} suggests that qualitative trends (which parameters cause failure) transfer well, even when quantitative thresholds differ.

\textbf{Task scope.} The current evaluation is limited to pick-and-place.
Extending to more complex manipulation tasks (\eg, insertion, assembly) requires additional scenario definitions and failure mode taxonomies.
However, the two-stage sampling methodology is task-agnostic and can be applied to any binary-outcome evaluation.

\textbf{Parameter independence.} LHS assumes that parameters can be varied independently, which may not hold for all real-world scenarios (\eg, mass and size are physically correlated for fixed-density objects).
The inclusion of interaction terms in the logistic regression model partially addresses this limitation, but correlated parameter distributions would require copula-based sampling strategies.

\textbf{Model simplicity.} Logistic regression captures linear and pairwise interaction effects but misses higher-order nonlinearities.
The AUC of 0.780 suggests room for improvement with more expressive models, at the cost of interpretability.
We deliberately prioritize interpretability for industrial deployment, where operators must understand and trust the risk model's predictions.

\textbf{Single-task evaluation.} Both robots are evaluated on the same pick-and-place task, limiting our ability to draw conclusions about task-dependent failure modes.
A multi-task evaluation campaign would reveal whether the identified danger zones are task-specific or generalize across manipulation primitives.

\subsection{Future Directions}

Several promising extensions of this work are worth highlighting:

\textbf{Active learning integration.}
The current two-stage approach uses a fixed Stage~1/Stage~2 split.
An active learning variant could iteratively allocate experiments to the most uncertain regions, potentially requiring fewer total experiments to achieve the same boundary resolution.
Gaussian process-based acquisition functions~\cite{chaloner1995bayesian} are a natural choice, though scalability to 8+ dimensions requires approximations such as random Fourier features.

\textbf{Multi-task evaluation.}
Extending \robogate{} to assembly, insertion, and tool-use tasks would enable cross-task failure analysis.
We hypothesize that some danger zones (high mass, low friction) are universal across manipulation primitives, while others (e.g., orientation sensitivity for insertion tasks) are task-specific.

\textbf{Real-world validation.}
A critical next step is validating the simulated failure boundaries against real-world experiments.
We plan to conduct a 500-experiment real-world campaign on a Franka Panda, sampling from both safe and danger zones as identified by the simulation study.
The primary research question is whether the boundary equation (Eq.~\ref{eq:boundary_equation}) transfers quantitatively or only qualitatively to the real robot.

\textbf{VLA model fine-tuning.}
The failure dictionary produced by \robogate{} can serve as targeted training data for VLA models.
By fine-tuning on the adversarial conditions where the model fails most severely (low lighting, clutter), we expect significant improvements in robustness.
A closed-loop workflow---evaluate with \robogate{}, fine-tune on failures, re-evaluate---could accelerate VLA model development for industrial deployment.

\textbf{Multi-robot fleet validation.}
Industrial deployments often involve fleets of identical robots that may exhibit unit-to-unit variation.
Extending \robogate{} to characterize the failure boundary distribution across a fleet (rather than a single robot) would provide fleet-level deployment confidence.

\section{Conclusion}
\label{sec:conclusion}

We presented \robogate, a deployment risk management framework for industrial robot policies that combines physics-based simulation with two-stage adaptive sampling to efficiently discover and characterize failure boundaries.
Our key findings from 40,000+ experiments across four robot platforms and seven VLA models are:

\begin{enumerate}[nosep,leftmargin=*]
\item Boundary-focused sampling improves risk model AUC from 0.754 to 0.780, demonstrating the value of concentrating experiments in the success-failure transition zone.

\item The failure boundary in friction-mass space follows a closed-form equation $\mu^*(m) = (1.469 + 0.419m)/(3.691 - 1.400m)$, enabling direct translation to operational constraints.

\item Cross-robot evaluation across four embodiments (Franka Panda, UR3e, UR5e, UR10e) reveals a non-monotonic reach-performance relationship: UR5e (0.85\,m, SR 74.3\%) and UR10e (1.30\,m, SR 61.6\%) outperform Franka Panda (0.855\,m, SR 33.3\%) and UR3e (0.50\,m, SR 9.6\%), demonstrating that suction grippers and reach-to-workspace ratio jointly determine deployment risk.

\item All seven VLA models achieve 0\% SR on \robogate{}'s 68 industrial scenarios vs.\ 100\% for the scripted controller.
Most critically, GR00T N1.6 fine-tuned to 97.65\% on LIBERO (MuJoCo) scores 0\% on Isaac Sim---a 97.65 percentage point cross-simulator gap that demonstrates academic benchmark performance does not predict deployment readiness.

\item The Confidence Score captures meaningful safety gradients even at 0\% SR: GR00T N1.6 (finetuned) scores 49/100 with zero collisions, while SmolVLA scores 1/100 with 66 collisions, providing actionable risk differentiation among uniformly unsuccessful models.
\end{enumerate}

These results argue for a \emph{validation layer} paradigm in Physical AI: just as quantum computing requires AI-based calibration layers to make noisy processors practical~\cite{nvidia_ising_2026}, robot foundation models require simulator-specific validation before deployment.
Future work will extend \robogate{} to multi-step manipulation tasks, incorporate active learning for boundary refinement, and validate the sim-to-real transfer of discovered failure boundaries.

\robogate{} is open-source at \url{https://github.com/liveplex-cpu/robogate}.
The full failure dictionary (30K+ experiments) is available at \url{https://huggingface.co/datasets/liveplex/robogate-failure-dictionary}.


\appendix

\section{Failure Dictionary Schema}
\label{app:schema}

Each experiment in the failure dictionary contains 26 fields (Franka) or 10 fields (UR5e).
Table~\ref{tab:schema_franka} documents the full Franka schema.

\begin{table}[h]
\centering
\small
\caption{Franka failure dictionary schema (26 fields per experiment).}
\label{tab:schema_franka}
\begin{tabular}{@{}lll@{}}
\toprule
\textbf{Field} & \textbf{Type} & \textbf{Description} \\
\midrule
\multicolumn{3}{@{}l}{\textit{Physical parameters}} \\
friction & float & Friction coefficient [0.05, 1.2] \\
mass & float & Object mass in kg [0.05, 2.0] \\
com\_offset & float & COM offset [0, 0.4] \\
size & float & Object size in m [0.02, 0.12] \\
ik\_noise & float & IK noise in rad [0, 0.04] \\
obstacles & int & Obstacle count [0, 5] \\
shape & str & box/cylinder/sphere/irregular \\
placement & str & Placement configuration \\
\midrule
\multicolumn{3}{@{}l}{\textit{Outcome fields}} \\
success & bool & Episode success \\
failure\_type & str & none/timeout/collision/... \\
cycle\_time & float & Episode duration in seconds \\
collision & bool & Collision occurred \\
drop & bool & Object dropped \\
grasp\_miss & bool & Grasp missed \\
\midrule
\multicolumn{3}{@{}l}{\textit{Derived fields}} \\
fail\_prob & float & Analytical failure probability \\
zone & str & safe/boundary/danger \\
sample\_idx & int & LHS sample index \\
\bottomrule
\end{tabular}
\end{table}

\section{Failure Mode Transition Details}
\label{app:transitions}

Table~\ref{tab:transitions} shows the dominant failure mode at each friction level, demonstrating the systematic transition from timeout-dominated failures at low friction to collision-dominated failures at high friction.

\begin{table}[h]
\centering
\small
\caption{Dominant failure mode by friction bin (Franka 20K). Transition from \texttt{timeout} to \texttt{collision} occurs at $\mu \approx 0.77$.}
\label{tab:transitions}
\begin{tabular}{@{}lclr@{}}
\toprule
\textbf{Friction Range} & \textbf{N fail} & \textbf{Dominant Mode} & \textbf{\%} \\
\midrule
0.050--0.108 & 1,968 & timeout & 39.1\% \\
0.108--0.165 & 1,069 & timeout & 44.3\% \\
0.165--0.280 & 1,298 & timeout & 42.0\% \\
0.280--0.450 & 1,514 & timeout & 38.5\% \\
0.450--0.625 & 1,067 & grip\_loss & 33.1\% \\
0.625--0.768 & 1,044 & timeout & 31.2\% \\
0.768--1.200 & 2,323 & collision & 42.8\% \\
\bottomrule
\end{tabular}
\end{table}

\section{UR5e Parameter Space}
\label{app:ur5e}

The UR5e evaluation uses a reduced 5-dimensional parameter space, reflecting the suction gripper's insensitivity to friction and shape parameters.

\begin{table}[h]
\centering
\small
\caption{UR5e parameter space (5 dimensions).}
\label{tab:ur5e_params}
\begin{tabular}{@{}llc@{}}
\toprule
\textbf{Parameter} & \textbf{Range} & \textbf{Scale} \\
\midrule
ik\_noise & [0, 0.02] rad & linear \\
mass & [0.05, 3.0] kg & log \\
grip\_threshold & [0.005, 0.02] & linear \\
obstacles & [0, 3] & integer \\
placement & 8 configurations & categorical \\
\bottomrule
\end{tabular}
\end{table}

\section{VLA Evaluation: Complete Per-Variant Results}
\label{app:vla_variants}

Table~\ref{tab:vla_variants} shows the full per-variant breakdown for the Octo-Small VLA evaluation, sorted by success rate (worst first).

\begin{table}[h]
\centering
\small
\caption{Octo-Small VLA: per-variant results on 68 scenarios. SR = success rate.}
\label{tab:vla_variants}
\begin{tabular}{@{}lcrrl@{}}
\toprule
\textbf{Category/Variant} & \textbf{Pass} & \textbf{Total} & \textbf{SR} & \textbf{Primary Fail} \\
\midrule
nom/standard\_objects & 0 & 7 & 0\% & grasp\_miss (6), collision (1) \\
nom/standard\_lighting & 0 & 7 & 0\% & grasp\_miss (5), collision (2) \\
nom/centered\_placement & 0 & 6 & 0\% & grasp\_miss (5), collision (1) \\
edge/small\_objects & 0 & 3 & 0\% & grasp\_miss (2), collision (1) \\
edge/heavy\_objects & 0 & 3 & 0\% & grasp\_miss (3) \\
edge/edge\_placement & 0 & 3 & 0\% & grasp\_miss (2), collision (1) \\
edge/occluded\_objects & 0 & 3 & 0\% & grasp\_miss (2), collision (1) \\
edge/transparent\_objects & 0 & 3 & 0\% & grasp\_miss (3) \\
adv/low\_lighting & 0 & 3 & 0\% & grasp\_miss (2), collision (1) \\
adv/cluttered\_scene & 0 & 3 & 0\% & grasp\_miss (3) \\
adv/slippery\_surface & 0 & 2 & 0\% & grasp\_miss (2) \\
adv/moving\_disturbance & 0 & 2 & 0\% & grasp\_miss (2) \\
dr/lighting & 0 & 10 & 0\% & grasp\_miss (7), collision (3) \\
dr/color & 0 & 5 & 0\% & grasp\_miss (3), collision (2) \\
dr/position & 0 & 5 & 0\% & grasp\_miss (5) \\
dr/camera & 0 & 3 & 0\% & grasp\_miss (2), collision (1) \\
\bottomrule
\end{tabular}
\end{table}

\section{Confidence Score Computation}
\label{app:confidence}

The \robogate{} Confidence Score is a weighted combination of five metric assessments, designed to produce a single deployment readiness number between 0 and 100.

\begin{equation}
C = \sum_{i=1}^{5} w_i \cdot s_i(\mathbf{m})
\end{equation}

where $w_i$ are the metric weights and $s_i(\mathbf{m})$ are the normalized scores for each metric:

\begin{table}[h]
\centering
\small
\caption{Confidence Score weight allocation and scoring functions.}
\label{tab:confidence_weights}
\begin{tabular}{@{}llcl@{}}
\toprule
\textbf{Component} & \textbf{Metric} & \textbf{Weight} & \textbf{Scoring} \\
\midrule
Grasp SR & $N_s/N$ & 0.30 & Linear: $\min(1, \text{SR}/0.92)$ \\
Cycle Time & $\bar{t}_c$ & 0.20 & $1 - |\Delta t / t_{\text{base}}|$ \\
Collision & $n_{\text{col}}$ & 0.25 & $1$ if $n=0$, else $0$ \\
Edge Cases & SR$_{\text{edge}}$ & 0.15 & Linear: SR on edge scenarios \\
Baseline $\Delta$ & $\Delta\text{SR}$ & 0.10 & $\max(0, 1 + \Delta\text{SR})$ \\
\bottomrule
\end{tabular}
\end{table}

The collision component is binary (0 or 25 points)---any collision results in zero contribution from this component, reflecting the zero-tolerance policy for safety-critical events.
This design means that a policy with even one collision cannot score above 75/100, regardless of other metrics.

For the VLA evaluation (Section~\ref{sec:vla}), Octo-Small scores 1/100 due to: SR component = $0.30 \times (0.0/0.92) \times 100 = 0.0$, collision component = $0.25 \times 0 = 0$ (14 collisions), cycle time component $\approx 0$ (71.7s vs.\ baseline 4.1s), edge case component = $0.15 \times 0 = 0$ (0\% edge SR), and baseline delta = $0.10 \times 0 = 0$ ($-100$pp regression). The minimum score of 1.0 reflects the worst possible outcome across all five metrics simultaneously.

\section{Boundary-Focused Sampling Algorithm Details}
\label{app:boundary_algo}

The boundary detection algorithm in Stage~2 operates as follows:

\begin{algorithm}[h]
\caption{Boundary Region Detection}
\label{alg:boundary_detect}
\begin{algorithmic}[1]
\Require Stage~1 experiments $\mathcal{D}_1$, SR range $[sr_l, sr_h]$
\Ensure Narrowed parameter ranges $\mathcal{R}_{\text{bnd}}$
\For{each continuous parameter $p \in \{$friction, mass, com\_offset, size, ik\_noise$\}$}
  \State Partition $p$ range into 10 equal-width bins $B_1, \ldots, B_{10}$
  \For{each bin $B_k$}
    \State $\text{SR}_k \gets$ success rate of experiments in $B_k$
  \EndFor
  \State $\mathcal{B}_p \gets \{B_k : sr_l \leq \text{SR}_k \leq sr_h\}$
  \If{$|\mathcal{B}_p| > 0$}
    \State $\mathcal{R}_{\text{bnd}}[p] \gets [\min(\mathcal{B}_p), \max(\mathcal{B}_p)]$
  \Else
    \State $\mathcal{R}_{\text{bnd}}[p] \gets$ fallback range from domain knowledge
  \EndIf
\EndFor
\State \textbf{return} $\mathcal{R}_{\text{bnd}}$
\end{algorithmic}
\end{algorithm}

The fallback ranges are derived from domain knowledge about typical industrial conditions:
\begin{itemize}[nosep,leftmargin=*]
\item Friction: [0.05, 0.40] --- most industrial objects have low-to-moderate friction
\item Mass: [0.3, 2.0] kg --- typical payload range for collaborative robots
\item COM offset: [0.1, 0.4] --- irregular objects with shifted center of mass
\item Size: [0.015, 0.05] m --- small parts that are difficult to grasp
\item IK noise: [0.01, 0.04] rad --- typical encoder uncertainty range
\end{itemize}

The emphasis region ($\mu < 0.3$, $m \geq 0.5$) receives 30\% of the Stage~2 budget because preliminary analysis showed the steepest SR gradient in this region (SR changes from 60\% to 10\% within a narrow friction range at moderate mass).

\section{Reproducibility}
\label{app:reproducibility}

All experiments are reproducible with the following commands.
Seeds are fixed to ensure deterministic results.

{\small
\begin{verbatim}
# Stage 1: Franka uniform (10K)
python scripts/generate_failure_dictionary_large.py \
  --n 10000 --seed 2026

# Stage 1: UR5e uniform (10K)
python scripts/test_ur5e_baseline.py \
  --n 10000 --seed 2026

# Stage 2: Franka boundary (10K)
python scripts/generate_boundary_focused.py \
  --n 10000 --seed 2024 \
  --input failure_dictionary_large.json

# Boundary analysis
python scripts/analyze_boundary_data.py \
  --original failure_dictionary_large.json \
  --boundary franka_boundary_10k.json \
  --ur5e ur5e_failure_dictionary.json

# VLA evaluation (two-process ZMQ pipeline)
# Terminal 1: Isaac Sim server (Python 3.11)
C:\IsaacSim\_build\windows-x86_64\release\python.bat \
  scripts/vla_isaac_server.py --mss-camera --port 5555
# Terminal 2: Octo client (conda octo, Python 3.10)
conda run -n octo python scripts/vla_octo_client.py \
  --model octo-small --port 5555 \
  --output robogate_demo/vla_octo_eval_real.json

# Paper figures
python scripts/generate_paper_figures.py
\end{verbatim}
}

Hardware: NVIDIA RTX 4090 (24~GB VRAM), Intel i9-13900K, 64~GB RAM.
Software: Ubuntu 22.04, NVIDIA Isaac Sim 5.1, Python 3.12, NumPy 2.x.

\end{document}